\documentclass{article}

\usepackage{arxiv}

\usepackage[utf8]{inputenc} 
\usepackage[T1]{fontenc}    
\usepackage{url}            
\usepackage{booktabs}       
\usepackage{amsfonts}       
\usepackage{nicefrac}       
\usepackage{orcidlink}
\usepackage{multirow}
\usepackage{microtype}      
\usepackage{lipsum}
\usepackage{graphicx}
\graphicspath{ {./images/} }
\RequirePackage{amsthm,amsmath,amsfonts,amssymb}

\RequirePackage{algorithm}
\RequirePackage[noend]{algpseudocode}

\RequirePackage[numbers]{natbib}
\RequirePackage{fontenc,inputenc,subcaption,graphicx,sectsty,titlesec,caption}
\RequirePackage{amsthm,amsmath,amssymb,animate,nccmath}

\RequirePackage{paralist}
\algnewcommand\REQUIRE{\State \textbf{Require:}}
\algnewcommand\ENSURE{\State \textbf{Ensure:}}
\algnewcommand\RETURN{\State \textbf{return}}

\theoremstyle{remark}
\newtheorem{Remark}{Remark}

\def\b1{\mbox{$\mathbf{1}$}}

\def\bX{\mbox{$\mathbf{X}$}}

\newcommand{\betavec}{\boldsymbol{\beta}}

\theoremstyle{plain}

\theoremstyle{definition}

\theoremstyle{remark}

\title{Distribution-Free Inference for LightGBM and GLM with Tweedie Loss}

\author{
Alokesh Manna \orcidlink{0009-0007-7958-5273} \\
Department of Statistics \\
University of Connecticut \\
\texttt{alokesh.manna@uconn.edu} \\
\And
Aditya Vikram Sett \\
Department of Statistics \\
University of Connecticut \\
\texttt{aditya.vikram.sett@uconn.edu} \\
\And
Dipak K. Dey \\
Department of Statistics \\
University of Connecticut \\
\texttt{dipak.dey@uconn.edu} \\
\And
Yuwen Gu \\
Department of Statistics \\
University of Connecticut \\
\texttt{yuwen.gu@uconn.edu} \\
\And
Elizabeth D. Schifano \\
Department of Statistics \\
University of Connecticut \\
\texttt{elizabeth.schifano@uconn.edu} \\
\And
Jichao He \\
The Travelers Companies, Inc. \\
Hartford, CT, USA \\
\texttt{jhe@travelers.com}
}

\begin{document}
\maketitle
\begin{abstract}
Prediction uncertainty quantification is a key research topic in recent years 
scientific and business problems. In insurance industries (\cite{parodi2023pricing}), assessing the range of possible claim costs for individual drivers improves premium pricing accuracy. It also enables insurers to manage risk more effectively by accounting for uncertainty in accident likelihood and severity. In the presence of covariates, a variety of 
regression-type models are often used for modeling insurance claims, 
ranging from relatively simple generalized linear models (GLMs) to regularized 
GLMs to gradient boosting models (GBMs). Conformal predictive inference has arisen as a popular distribution-free approach for quantifying predictive uncertainty under relatively weak assumptions of exchangeability, and has been well studied under the classic linear regression setting. In this work, we propose new non-conformity measures for GLMs and GBMs with GLM-type loss. Using regularized Tweedie GLM regression and LightGBM with Tweedie loss, we demonstrate conformal prediction performance with these non-conformity measures in insurance claims data. Our simulation results favor the use of locally weighted Pearson residuals for LightGBM over other methods considered, as the resulting intervals maintained the nominal coverage with the smallest average width.  \\ 
\vspace{0.1in}
\end{abstract}	

\section{Introduction}

Quantifying prediction uncertainty is important in many applications \citep[e.g.,][]{zou2009simple, mudelsee2007quantifying, manna2024development}. If the data structure is high-dimensional and the sample size is low, additional challenges arise that require attention \citep{dawn2025some}. Recent advances have explored the integration of external data sources and uncertainty quantification in forecasting. For instance, \cite{aaronson2022forecasting} demonstrated how Google Trends can enhance real-time prediction of unemployment claims while accounting for out-of-sample uncertainty. Similarly, \cite{koissi2006evaluating} introduced a residual bootstrap approach to construct confidence intervals in mortality forecasting. Beyond linear model exploration, methods for prediction uncertainty quantification in generalized linear models (GLMs) and related regularized models remain limited \citep[see, e.g.,][]{woods2006designs, leyk2004predictive, dormann2008components}. The Tweedie distribution is a special and versatile member of the exponential dispersion family that offers unique modeling flexibility in GLMs, making it particularly valuable in applied fields such as insurance \citep{manna2024interval}, ecology \citep{hasan2011two}, and health economics \citep{mallick2022differential, kurz2017tweedie}. It is especially well-suited for datasets that exhibit a combination of right-skewed continuous outcomes, excess zeros, heavy tails, and non-constant variance—features that are often challenging for traditional distributions to accommodate simultaneously. These properties arise naturally when the Tweedie index parameter ``$p$'' lies between 1 and 2, corresponding to a compound Poisson-Gamma structure. As a result, the Tweedie distribution has gained popularity in actuarial science and GLM, where it is widely used for modeling insurance claim amounts, rainfall data, and healthcare costs, providing a coherent framework to handle semi-continuous outcomes with heterogeneous variability.

Quantifying uncertainty in predictions is particularly crucial for insurance regression models, especially when modeling auto claim data using Tweedie distributions. Auto insurance claims exhibit significant variability due to factors such as claim frequency, severity, and the presence of zero claims for policyholders with no accidents. 
This uncertainty quantification becomes essential for several reasons: first, it enables insurers to set appropriate premium reserves and confidence intervals around expected losses; second, it helps identify high-risk policies where prediction uncertainty is greatest, allowing for more informed underwriting decisions; and third, it supports regulatory compliance by providing transparent measures of model reliability. In the context of auto claim modeling, where claim amounts can range from zero to extremely large values, understanding the prediction uncertainty helps insurers distinguish between policies with genuinely low expected claims versus those with high uncertainty due to limited data or complex risk profiles. Without proper uncertainty quantification, insurers may underestimate their exposure to tail risks and fail to price policies appropriately. Traditional Tweedie regression models provide point estimates of expected claim costs along with prediction intervals, but these intervals tend to be quite wide due to the over-dispersion inherent in insurance data. The wide intervals can limit their practical utility for precise risk assessment and decision-making in insurance applications.

Conformal predictive inference has arisen as a promising approach for quantifying 
uncertainty around predictions, offering predictive intervals that are both 
model-agnostic and distribution-free under the mild assumption of exchangeability. 
It was formalized in the work of \cite{vovk2005algorithmic}, and it has grown 
as an essential tool in distribution-free predictive inference. A good tutorial 
of this resource can also be found in \cite{shafer2008tutorial}.
\cite{lei2018distribution} developed split conformal prediction for regression 
models, which separates the fitting and nonconformity ranking steps.
Although split conformal intervals tend to be wider than full conformal intervals,
the split conformal approach has nonetheless gained considerable traction due to 
its substantially reduced computational cost. 

While both conformal prediction and bootstrap methods aim to quantify uncertainty, they differ in foundational assumptions, implementation, guarantees, and in many applications, conformal methods offer several 
advantages. Conformal prediction guarantees marginal coverage (e.g., a 90\% interval will truly contain the response 90\% of the time theoretically) without assuming the model is correct as long as the non-conformity measures
 are exchangeable, 
 which is a milder assumption than independent and identically distributed.  In supervised learning, conformal prediction works for any estimator of the regression function (GLMs, GBMs, neural networks, etc). Among different existing methods, bootstrap intervals rely on asymptotic properties and often undercover in small or imbalanced samples \citep[see, e.g.,][]{bradley1993introduction}.
Conformal prediction does not require specifying the noise distribution or model likelihood \citep{lei2018distribution}; the bootstrap, in contrast, assumes resampled residuals adequately approximate the true error distribution. This is particularly important in Tweedie models, where claims may be zero-inflated and skewed, which makes residual bootstrapping less reliable or have larger intervals \citep{tibbe2022correcting}.
Conformal prediction methods have been explored in the context of insurance data by \cite{hong2021valid} and \cite{hong2023conformal}. 
While \cite{hong2023conformal} proposed a method that converts the credibility 
estimator into a conformal prediction credibility interval, \cite{hong2021valid} 
proposed a valid model-free prediction interval for insurance claims based on 
conformal prediction. \cite{hong2021valid} focused on a setting without covariates and with full (not split) conformal inference, but noted that the conformal approach could be extended to regression
models with a suitable non-conformity (residual) measure. 

In this work, we more closely examine the regression setting in the context
of high-dimensional GLMs and gradient boosting 
machine models (GBMs) in split conformal prediction.  More specifically, we examine several non-conformity measures tailored to the GLM framework, allowing for the construction of valid prediction intervals of claim amount that reflect 
customer-specific features and behaviors. We compare the performance of the 
different non-conformity measures in simulation using the Tweedie GLM, which 
generalizes the compound Poisson regression relevant to the insurance data, and 
also LightGBM using the Tweedie loss, which enables a complicated non-linear 
structure for covariates and target variables. 

The sections of this manuscript are organized as 
follows. We first review full and split conformal prediction as well as local weighting approaches, 
and then extend these ideas to define non-conformity measures potentially more appropriate for 
predictions obtained from GLM-type models in Section~\ref{methods}. We also include here a variant of locally weighted Pearson residuals for LightGBM. 
In Section~\ref{simulation}, we conduct a simulation study based on insurance auto claims data to compare the resulting split conformal prediction intervals in terms of both interval coverage and width. A discussion concludes in Section~\ref{discussion}.  The R code is 
available on GitHub for reproducibility: \url{https://github.com/alokesh17/conformal_LightGBM_tweedie.git}.

\section{Methods}\label{methods}

In this section, we examine different ways of obtaining conformal prediction intervals for a target based on a fitted GLM or LightGBM regression model.  More 
specifically, we focus on the Tweedie GLM and LightGBM with Tweedie loss, as both  
are commonly used in insurance modeling \citep[e.g.,][]{so2024enhanced, halder2021spatial, yang2018insurance, bortoluzzo2011estimating}. Our 
goal is to develop a prediction interval procedure that does not rely on strong 
parametric assumptions of the model being fit. More precisely, 
we seek an approach that remains valid even if the predictive 
model is misspecified. Among the available methods in the literature, conformal 
prediction offers a principled, distribution-free framework for constructing 
prediction intervals with finite-sample coverage guarantees.
We first briefly review full and split conformal predictive intervals for 
general regression problems, and an extension for handling heteroscedasticity.
We then discuss our proposed non-conformity measures tailored for predictions obtained 
from GLM-type models.

\subsection{Brief Review of Conformal and Split Conformal Prediction for General Regression Problems}

\begin{algorithm}[!t]
\caption{Conformal Prediction Interval Construction without Splitting \citep[adapted from ][]{lei2018distribution}.}
\label{alg:full}
\renewcommand{\algorithmicrequire}{\textbf{Input:}}
\renewcommand{\algorithmicensure}{\textbf{Output:}}
\begin{algorithmic}[1]
\REQUIRE Data $(X_i; Y_i)$, $i = 1,\ldots,n,$ miscoverage level $\alpha\in (0, 1)$, 
regression algorithm $\mathcal{A}$, points $\mathcal{X}_{new} = \{X_{n+1};X_{n+2}, \ldots \}$ 
at which to construct prediction intervals, and values $\mathcal{Y}_{trial} = \{ y_1, y_2, \ldots \}$ 
to act as trial values.
\ENSURE Predictions intervals, at each element of $\mathcal{X}_{new}$
\For{$x\in \mathcal{X}_{new}$}
  \For{$y\in \mathcal{Y}_{trial}$}
  \State Train on augmented data: Obtain function $\hat{f}(\cdot)$ using regression algorithm $\mathcal{A}$ 
	to estimate $\mathrm{E}(Y|X)$ using $\{(X_1; Y_1), \ldots, (X_n; Y_n), (x; y)\}$.
  \State Compute non-conformity measure (absolute residuals) 
	$R_{i}:=\big|Y_{i}- \hat{f}(X_{i}) \big|$ for $i = 1,\dots,n$ 	and $R_{n+1}=\big|y- \hat{f}(x) \big|.$
	\State Obtain proportion of points whose non-conformity measure is smaller than $R_{n+1}$: 
	$$\pi(y) =(n+1)^{-1}\sum_{i=1}^{n+1} I(R_i \leq R_{n+1}),$$
	where $I(a)$ is the indicator function equaling 1 when condition $a$ is true.
	\EndFor
\State Compute prediction interval for $x$: 
$C(x) = \{ y \in \mathcal{Y}_{trial}: (n+1)\pi(y) \leq \lceil(1-\alpha)(n+1)\rceil\}$
\EndFor
\State Return prediction interval $C(x)$ for each $x\in\mathcal{X}_{new}$.
\end{algorithmic}
\end{algorithm}

Conformal analysis is an evolving field in statistics, and its suitable applications make it an increasingly important area of focus. The works by \cite{vovk2005algorithmic}, \cite{lei2014distribution}, \cite{lei2018distribution} constitute foundational literature in conformal prediction, where researchers started to investigate its importance in the regression framework. Following \cite{lei2018distribution}, 
suppose we observe i.i.d. samples $Z_i = (X_i; Y_i)\in \mathcal{R}^d \times \mathcal{R}\sim P$,
$i = 1, \ldots, n,$ and we wish to construct a prediction interval for $Y_{n+1}, Y_{n+2}, \ldots,$ 
at new feature value(s) $X_{n+1}, X_{n+2}, \ldots$, where $(X_{n+1}, Y_{n+1})$, 
$(X_{n+2}, Y_{n+2}), \ldots$, are independent draws from $P$.  The procedure is 
summarized in Algorithm~\ref{alg:full}. The key observation is that by 
exchangeability of the data points and the assumed symmetry of $\hat{f}$, when 
evaluated at $y = Y_{n+1}$, we see that $\pi(Y_{n+1})$ (defined in Algorithm~\ref{alg:full}) is uniformly distributed 
over the set $\{1/(n + 1), 2/(n + 1), \ldots, 1\}$, which
implies
\[
\Pr\left( (n+1)\pi(Y_{n+1}) \leq \lceil(1-\alpha)(n+1)\rceil \right) \geq 1-\alpha.
\]
This further implies 
$C(x) = \{ y \in \mathcal{R}: (n+1)\pi(y) \leq \lceil(1-\alpha)(n+1)\rceil\},$
although in practice, we must use a discrete grid of trial values $y$.  Also
notice that the steps in the inner for loop must be repeated for each new feature 
value for which we want a prediction, which can be time consuming for more
complicated models in the training step.  This motivated the split conformal
prediction interval approach of \cite{lei2018distribution}, described in 
Algorithm~\ref{alg:split}.

The major difference is that in split conformal, we randomly split the $n$ 
observations into two subsets, with indices collected in $\mathcal{D}_{1}$ 
and $\mathcal{D}_{2}$, respectively, where $|\mathcal{D}_{1}|=n_1$ and 
$|\mathcal{D}_{2}|=n_2$. We train the model on the $n_1$ observations in 
$\mathcal{D}_{1}$ and then compute and rank the non-conformity 
measure on the separate calibration set $\mathcal{D}_{2}$. 
Based on the ranking of the non-conformity measures within the calibration set, 
we can then calculate the desired prediction interval.
\cite{lei2018distribution} showed that for any new i.i.d. draw $(X_{n+1}; Y_{n+1})$,
$\Pr(Y_{n+1}\in C_S(X_{n+1})) \geq 1-\alpha.$ The crucial observation is that the 
absolute residuals based on $\mathcal{D}_{2}$ are still exchangeable.
If we additionally assume that $R_i$, $i\in \mathcal{D}_{2},$ have a 
continuous joint distribution, then \cite{lei2018distribution} further showed
\[
\Pr(Y_{n+1} \in C_S(X_{n+1})) \in \Bigl[1-\alpha, 1-\alpha+\frac{1}{n_{2}+1}\Bigr).
\]

\cite{lei2018distribution} suggest using an even split to create $\mathcal{D}_{1}$ 
and $\mathcal{D}_{2}$, i.e., $n_1=n_2=n/2$ for even 
sample size $n$, but noted split conformal inference can also be implemented using 
an unbalanced split. Particularly in situations where the regression
procedure is complex, it may be beneficial to have $n_1>n_2$ so that the trained
estimator is more accurate.  Like \cite{lei2018distribution}, we focus on the
equal split in our empirical analyses.

\begin{algorithm}[!t]
\caption{Prediction Interval Construction with Data Splitting \citep[adapted from ][]{lei2018distribution}.}
\label{alg:split}
\renewcommand{\algorithmicrequire}{\textbf{Input:}}
\renewcommand{\algorithmicensure}{\textbf{Output:}}
\begin{algorithmic}[1]
\REQUIRE Data $(X_i, Y_i)$ for $i = 1,\ldots,n$, miscoverage level $\alpha \in (0, 1)$, 
regression algorithm $\mathcal{A}$.
\ENSURE Prediction interval for $x \in \mathbb{R}^d$.
\State Randomly split indices $i = 1, \ldots, n$ into two subsets, $\mathcal{D}_1$ and 
$\mathcal{D}_2$, with $|\mathcal{D}_1| = n_1$ and $|\mathcal{D}_2| = n_2$.
\State \textbf{Train on $\mathcal{D}_1$:} Use $\mathcal{A}$ to fit regression function $\hat{f}_{n_1}(x)$ to data in $\mathcal{D}_1$.
\State\textbf{Compute residuals on $\mathcal{D}_2$:} For each $i \in \mathcal{D}_2$, compute unstandardized residuals $R_i = |Y_i - \hat{f}_{n_1}(X_i)|$.
\State \textbf{Compute nonconformity quantile $q_{n_2}$:} Let $q_{n_2}$ be the $k$th smallest value in $\{R_i : i \in \mathcal{D}_2\}$, where $k = \lceil(n_2 + 1)(1 - \alpha)\rceil$.
\State \textbf{Return prediction interval:} 
\[
C_S(x) = \left[ \hat{f}_{n_1}(x) - q_{n_2},\; \hat{f}_{n_1}(x) + q_{n_2} \right], \quad \text{for } x \in \mathbb{R}^d.
\]
\end{algorithmic}
\end{algorithm}

\cite{lei2018distribution} also proposed a variant of Algorithm~\ref{alg:split}
described as \textit{local weighting} which can help in cases of heteroscedasticity. 
This involves replacing $R_i$ in Algorithm~\ref{alg:split} with standardized 
absolute residuals
\[
R_{i} = \frac{|Y_{i} - \hat{f}_{n_{1}}(X_{i})|}{\hat{\sigma}(X_{i})},~i\in\mathcal{D}_2,
\]
where $\hat{\sigma}(X_{i})$ is an estimate of the error spread.  \cite{lei2018distribution} set  $\hat{\sigma}(X_i)$ as an estimate of the conditional mean absolute deviation (MAD) of $|Y - f(X)| |X=x$, as a function of $x \in\mathcal{R}^d$ based on the fitted residuals obtained 
from $\mathcal{D}_1$.
The prediction interval is then modified accordingly as 
\[
C_S(x) = \left[ \hat{f}_{n_{1}}(x) - \hat{\sigma}(x){q}_{n_{2}},\; \hat{f}_{n_{1}}(x) + \hat{\sigma}(x){q}_{n_{2}} \right],
\]
where ${q}_{n_{2}}$ is the $\lceil{(1-\alpha)(n_{2}+1)}\rceil$th 
smallest value of the standardized absolute residuals.

\subsection{Split Conformal Prediction and GLM}

GLMs are flexible generalizations of the ordinary linear (least squares) regression
that allow for different response distributions from the exponential dispersion
family and link functions that connect the mean function to the linear predictor.  See, e.g., 
\cite{mccullagh89_genera_linear_models} for complete details.  While the conformal
approaches discussed above are indeed model-agnostic, there are many types of residuals
(and hence non-conformity measures) one could consider after fitting a GLM or GLM-type
model (e.g., regularized GLM for variable selection or LightGBM with GLM-based 
loss).  We consider below three commonly used GLM residuals as potential 
non-conformity measures, and focus on the Tweedie compound Poisson-Gamma specification due to its popularity in insurance modeling \citep[e.g., ][]{bonat2018extended, abid2023choice}.

For a response $Y$ belonging to the exponential dispersion family, its 
probability density function may be expressed as
  \begin{equation*}
    p_{Y}(y;\theta,\phi)=\exp\left\{\frac{y\theta-b(\theta)}{a(\phi)}
      +c(y,\phi)\right\},
  \end{equation*}
where $\theta$ is the canonical parameter and $\phi>0$ is the dispersion parameter.
Let $b'(\theta)$ and $b''(\theta)$ be respectively the first and second
derivatives of $b(\theta)$. The mean and variance of $Y$ can be expressed as
    \[
      \mu := \text{E}(Y)=b'(\theta), ~\text{var}(Y)=b''(\theta)a(\phi)=V(\mu)a(\phi),
    \] 
where $V(\mu)=b''\{(b')^{-1}(\mu)\}$ is called the variance function.  
Covariates in GLM are incorporated through the mean as 
$\mu = \text{E}(Y|X=x) = g^{-1}(x^T\beta)$ where $\beta$ is a $d$
dimensional vector and $g(\cdot)$ is the link function.

With data $(X_i,Y_i),~i=1,\ldots,n$, fit to a standard unregularized GLM,
we obtain the maximum likelihood estimate (MLE) $\widehat{\mu}_i= g^{-1}(X_i^T\hat{\beta}),$ 
where $\hat{\beta}$ is the MLE of the $d$-dimensional coefficient vector.  
Arguably, the three most common GLM residuals used for model fit diagnostics are the Pearson residuals, deviance residuals, and Anscombe residuals 
\citep[see, e.g., ][chapters 2 and 4]{mccullagh89_genera_linear_models}. 
See Table~\ref{tab:res} for general definitions of these residuals, as well as pros and cons of each. 


With an eye towards conformal inference, these three types of residuals are presumed more appropriate than $|Y_i -\widehat{\mu}_i|$ 
as they account for heteroscedasticity and therefore are more likely to satisfy the exchangeability assumption.  For split conformal prediction
in particular, we can use the absolute values of these 
residuals in a similar way as the variant of Algorithm~\ref{alg:split} that incorporates local weighting to accommodate heteroscedasticity.  An advantage
of the conformal approach is that we do not need to use the MLE 
$\widehat{\mu_i}= g^{-1}(X_i^T\hat{\beta})$, but rather other estimates
of the mean function $\hat{f}(X_i)$ obtained from other (and possibly mis-specified) 
GLM-type models (e.g, regularized GLM or LightGBM with GLM-type loss, as indicated earlier).

\begin{table}[!t]
\centering
\caption{Comparison of Pearson, Deviance, and Anscombe Residuals}
\label{tab:res}
\begin{tabular}{|c|p{3.7cm}|p{3.4cm}|p{3cm}|}
\hline
\textbf{Residual Type} & \textbf{Formula} & \textbf{Pros} &  \textbf{Cons}\\ \hline

\multirow{2}{*}{Pearson} &
$\frac{y_i - \hat{\mu}_i}{\sqrt{V(\hat{\mu}_i)}}$ , where $V(\hat{\mu}_i)$ is the variance function.

&
General diagnostics; simple to calculate; identifies overall model fit and heteroscedasticity. & Sensitive to model misspecification; not ideal for non-normal distributions. \\ \hline

\multirow{2}{*}{Deviance} &
$\text{sign}(y_i - \hat{\mu}_i) \sqrt{d_i}$ where,
$
d_i = 2 \left[ \ell(y_i \mid y_i) - \ell(y_i \mid \hat{\mu}_i) \right].
$&
Useful for detecting outliers and poor fits. Measures each observation's contribution to the model deviance.& Can be overly sensitive to outliers and large values. \\ \hline

\multirow{3}{*}{Anscombe} &
$
\frac{A(y_i) - A(\hat{\mu}_i)}{\sqrt{V(\hat{\mu}_i)}}$ where $A(.)$ is the variance stabilizing function $\int \frac{1}{\sqrt{V(\mu)}} \, d\mu$.
&
Stabilizes variance in non-normal data (e.g., Poisson or binomial). Approximates normality. & It is complex to calculate and not always easy to interpret. \\ \hline
\end{tabular}
\end{table}

\paragraph{Illustration with Tweedie Compound Poisson-Gamma Distribution}
\label{tweedie_start}
The Tweedie distribution is part of the exponential disperson family, and 
encompasses a wide range of well-known distributions depending on the value of 
the power parameter $p$.
We focus on the Tweedie compound Poisson-Gamma model, which corresponds to a 
Tweedie distribution with power parameter $p$ between 1 and 2.  
In exponential dispersion family form, the Tweedie model for $Y$ takes the following form:
\begin{equation*}
  p_{Y}(y|\mu,\phi,p)=\exp\left\{\frac{1}{\phi}\left(
      \frac{y\mu^{1-p}}{1-p}-\frac{\mu^{2-p}}{2-p}\right)
    +c(y,\phi,p)\right\},
\end{equation*}
where $\phi>0$, \(p\in(1,2)\), \(a=(2-p)/(p-1)\),  and
\begin{equation*}
  c(y,\phi,p)=\left\{
    \begin{array}{ll}
      0, & y = 0\\
      \log\left[\frac1y\sum_{j=1}^{\infty}\frac{y^{j \times a}}{\phi^{j(1+a)}(2-p)^{j}(p-1)^{j \times a}j!\Gamma(j \times a)}\right], & y>0.
    \end{array}
  \right.
\end{equation*}
In the application of insurance claims, the Tweedie distribution is a well-known 
distribution in the literature for handling datasets characterized by a mixture 
of zeros and positive continuous values. This distribution is particularly 
effective in modeling claim amounts, where many observations may be zero 
(e.g., no claims) and non-zero values are highly skewed. The Tweedie compound 
Poisson-Gamma distribution captures these features by combining a Poisson 
distribution for the count of events and a Gamma distribution for the magnitude 
of the events, making it an ideal choice for actuarial and insurance analytics. 
With the natural logarithm as the GLM link function to covariates $X$, the mean and variance 
are respectively $\mu=\text{E}(Y|X=x)=\exp(x^{T}\beta)$ and $\text{var}(Y|X=x)=\phi\mu^{p}$, 
where $V(\mu) = \mu^{p}.$

With a focus on split conformal prediction fit using Tweedie-type models, we 
consider the following non-conformity measures and resulting intervals.  
\begin{itemize}
\item Absolute Pearson residuals: 
\begin{equation*}
    R_{i,\text{Pear}}= \left| \frac{Y_{i}-\hat{f}_{n_1}(X_{i})}
    {\widehat{\phi}^{1/2}[\hat{f}_{n_1}(X_{i})]^{p/2}} \right|,~i\in\mathcal{D}_2,
\end{equation*}		
where $\hat{f}_{n_1}$ is an estimate of $\text{E}(Y|X)$ based on training set \(\mathcal{D}_{1}\).  
Note that the dispersion estimate \(\widehat{\phi}\) should also be obtained using training set \(\mathcal{D}_{1}\).  Since we assume $\phi$ is constant for
	all observations, \(\widehat{\phi}\) does not play a role in constructing
  the conformal prediction interval as it appears in all \(R_{i,\text{Pear}}\), $i\in\mathcal{D}_{2}\cup\{n+1\},$ and will cancel out in the resulting 
	interval. Thus, we instead use
  \begin{equation}\label{eq:resp}
    R_{i, \text{Pear}}=\left| \frac{Y_{i}-\widehat{f}_{n_{1}}(X_{i})}
    {[\widehat{f}_{n_{1}}(X_{i})]^{p/2}} \right|,~
    i\in\mathcal{D}_{2}.
  \end{equation}
Let $q_{n_2, \text{Pear}}$ be the \(\lceil{(1-\alpha)(n_{2}+1)}\rceil\)th smallest of \(R_{i, \text{Pear}},~i\in\mathcal{D}_{2}\),
The Pearson residual is the easiest to use for constructing the prediction
interval for \(Y_{n+1}\), with 
\begin{equation*}
  \left[\widehat{f}_{n_{1}}(X_{n+1})-{q}_{n_{2},\text{Pear}}\cdot[\widehat{f}_{n_{1}}(X_{n+1})]^{p/2},
  \widehat{f}_{n_{1}}(X_{n+1})+{q}_{n_{2},\text{Pear}}\cdot[\widehat{f}_{n_{1}}(X_{n+1})]^{p/2}\right].
\end{equation*}
A left truncation at zero may be needed if the lower end falls below zero. 
Thus, the interval we use in practice is
\begin{align}
C_{S, \text{Pear}}(X_{n+1}) = 
\Big[ & \left( \widehat{f}_{n_1}(X_{n+1}) 
- {q}_{n_2, \text{Pear}} \cdot \left[\widehat{f}_{n_1}(X_{n+1})\right]^{p/2} \right) \vee 0,\notag \\
& \widehat{f}_{n_1}(X_{n+1}) 
+ {q}_{n_2, \text{Pear}} \cdot \left[\widehat{f}_{n_1}(X_{n+1})\right]^{p/2} \Big].
\label{eq:intp}
\end{align}

\item Absolute Deviance residuals:
  \begin{equation}\label{eq:resd}
    R_{i, \text{Dev}}= 
    \sqrt{2\left(\frac{Y_{i}[\widehat{f}_{n_{1}}(X_{i})]^{1-p}}{p-1}
      -\frac{Y_{i}^{2-p}}{(p-1)(2-p)}
      +\frac{[\widehat{f}_{n_{1}}(X_{i})]^{2-p}}{2-p}\right)},
    ~i\in\mathcal{D}_{2}.
  \end{equation}
Let $q_{n_2, \text{Dev}}$ be the \(\lceil{(1-\alpha)(n_{2}+1)}\rceil\)th smallest of 
\(R_{i, \text{Dev}},i\in\mathcal{D}_{2}\),
For the deviance residual, solving for \(Y_{n+1}\) in order to construct the 
interval is a bit challenging. In practice, we employ a root-finding procedure 
to calculate $C_{S,\text{Dev}}(X_{n+1})$.

	
\item Absolute Anscombe residuals:
  \begin{equation*}
    R_{i, \text{Ansc}}= \left| \frac{\frac{3}{3-p}(Y_{i}^{1-p/3}-[\widehat{f}_{n_{1}}(\bX_{i})]^{1-p/3})}
    {[\widehat{f}_{n_{1}}(\bX_{i})]^{p/6}}\right|, ~i\in\mathcal{D}_{2}.
  \end{equation*}
  Similarly, the factor \(3/(3-p)\) does not matter as a constant, and will get 
	canceled out during interval construction. 
	Therefore, in practice we use
  \begin{equation}\label{eq:resa}
    R_{i, \text{Ansc}}=\left|\frac{Y_{i}^{1-p/3}-[\widehat{f}_{n_{1}}(X_{i})]^{1-p/3}}
    {[\widehat{f}_{n_{1}}(X_{i})]^{p/6}}\right|,~i\in\mathcal{D}_{2}.
  \end{equation}
Let $q_{n_2, \text{Ansc}}$ be the \(\lceil{(1-\alpha)(n_{2}+1)\rceil}\)th smallest of 
\(R_{i, \text{Ansc}},~i\in\mathcal{D}_{2}\).  
We obtain the following interval for \(Y_{n+1}\):
\begin{equation*}
   \begin{split}
    \biggl[\Bigl([\widehat{f}_{n_{1}}(X_{n+1})]^{1-p/3}
    &-{q}_{n_{2},\text{Ansc}}\cdot[\widehat{f}_{n_{1}}(X_{n+1})]^{p/6}
    \Bigr)^{3/(3-p)},\\
    &\Bigl([\widehat{f}_{n_{1}}(\bX_{n+1})]^{1-p/3}
      +{q}_{n_{2},\text{Ansc}}\cdot[\widehat{f}_{n_{1}}(X_{n+1})]^{p/6}
      \Bigr)^{3/(3-p)}\biggr].
  \end{split}
\end{equation*}
Again, a left truncation at zero may be needed when the lower end falls below
zero. Thus, in practice we take
\begin{align}
C_{S,\text{Ansc}}(X_{n+1}) = \Big[ &
\left\{ \left( [\widehat{f}_{n_1}(\mathbf{X}_{n+1})]^{1 - p/3}
- {q}_{n_2, \text{Ansc}} \cdot [\widehat{f}_{n_1}(\mathbf{X}_{n+1})]^{p/6} \right) \vee 0 \right\}^{3/(3 - p)}, \notag \\
&
\left( [\widehat{f}_{n_1}(\mathbf{X}_{n+1})]^{1 - p/3}
+ q_{n_2,\text{Ansc}} \cdot [\widehat{f}_{n_1}(\mathbf{X}_{n+1})]^{p/6} \right)^{3/(3 - p)} \Big].
\label{eq:inta}
\end{align}
\end{itemize}
The interval construction for the three residuals is summarized in 
Algorithm~\ref{alg:glm}.

\begin{algorithm}[!t]
\caption{Prediction Interval Construction with Data Splitting for Tweedie-type
Fitted Models}
\label{alg:glm}
\renewcommand{\algorithmicrequire}{\textbf{Input:}}
\renewcommand{\algorithmicensure}{\textbf{Output:}}
\begin{algorithmic}[1]
\REQUIRE Data $(X_i; Y_i)$, $i = 1,\ldots,n,$ miscoverage level $\alpha\in (0, 1)$, 
Tweedie-type regression algorithm $\mathcal{A}$, residual type $res\in\{\text{Pear},~\text{Dev},~\text{Ans}\}$.
\ENSURE Prediction band for $x\in\mathcal{R}^d$.
\State Randomly split $i=1,\ldots,n$ into two subsets, $\mathcal{D}_{1}$ and 
$\mathcal{D}_{2}$, where $|\mathcal{D}_{1}|=n_1$ and $|\mathcal{D}_{2}|=n_2.$
\State Train on $\mathcal{D}_{1}$: Obtain estimate $\hat{f}_{n_1}(x)$ of $\mathrm{E}(Y|X=x)$
using regression algorithm $\mathcal{A}$ on observations from $\mathcal{D}_{1}$.
\State Compute non-conformity measure (absolute residuals) using $\mathcal{D}_{2}$: 
Use $R_{i,res}$, as defined in Equation~\eqref{eq:resp}, \eqref{eq:resd}, or~\eqref{eq:resa}, 
for $i \in \mathcal{D}_2$.
\State Compute non-conformity quantile $q_{n_2,res}$: $q_{n_2, res}$ = the $k$th smallest 
value in $\{R_{i,res}, i\in\mathcal{D}_2\}$, where $k = \lceil(n_2 + 1)(1 - \alpha)\rceil$. 
\State Return prediction set 
$C_{S,res}(x)$ according to Equation~\eqref{eq:intp} for Pearson, a root-finding procedure for Deviance residuals (no closed form), or~\eqref{eq:inta} for $x\in\mathcal{R}^d$ for Anscombe residuals.
\end{algorithmic}
\end{algorithm}

\begin{Remark}
The intervals above are considered \textit{symmetric} intervals, as we add
and subtract the same value based on the same quantile $q_{n_2, res}$ to form the interval, where  $res \in \{\text{Pear},~\text{Dev},~\text{Ansc}\}.$  Due to the skewed 
and asymmetric nature of Tweedie compound Poisson-Gamma distribution (to match the skewed and asymmetric nature of many types of insurance data), it is possible 
that the standardized residuals may also have an asymmetric distribution.  Thus, 
we also considered \textit{asymmetric} intervals by defining two different 
quantiles, $q_{n_2, L, res}$ and $q_{n_2, U, res}$ for $res \in \{\text{Pear},~\text{Dev},~\text{Ansc}\},$ 
by selecting the \(\lfloor{(\alpha/2)(n_{2}+1)\rfloor}\)th and 
\(\lceil{(1-\alpha/2)(n_{2}+1)\rceil}\)th smallest raw residuals 
(i.e., without absolute value), respectively, from 
$\mathcal{D}_{2}$. We found in the simulation, however, that while these asymmetric intervals did have the correct coverage, they tended to have much longer widths than the corresponding symmetric intervals for all three types of residuals considered (see Supplemental Materials Section 1). 
\end{Remark}

\subsection{Locally Weighted Pearson Residuals}
\label{localized_formulation}
In the spirit of the locally weighted method inspired by \cite{lei2018distribution}, we propose a variant of the locally weighted method for Pearson residuals.  As demonstrated below in Section~\ref{simulation}, the prediction intervals based on  Pearson residuals outperformed those based on Deviance and Anscombe residuals, so we focus on a locally weighted variant for the Pearson residual only. Locally weighted Deviance and Anscombe residuals could be developed using a similar procedure. 

Recall the absolute Pearson residuals $R_{i, Pear},$ $i\in\mathcal{D}_2$, given in Equation~\eqref{eq:resp}.
We propose to locally weight these absolute Pearson residuals using
\begin{equation}
\label{eq:respw}
R^{*}_{i, \text{Pear}}=\frac{R_{i, \text{Pear}}}{\hat{\rho}_{n_1}(X_{i})},~
i\in\mathcal{D}_{2},
\end{equation}
where $\hat{\rho}_{n_1}(X_{i})$ is an estimate of the error spread. To estimate $\rho_{n_1}(X_{i})$, we regress the residuals $R_{i, \text{Pear}}$ with the covariates $X_{i}$ using a second model, a LightGBM model with squared-error loss, 
for $i\in\mathcal{D}_{1}$.  Let $q^{*}_{n_2, \text{Pear}}$ be the \(\lceil{(1-\alpha)(n_{2}+1)}\rceil\)th smallest of \(R^{*}_{i, Pear},~i\in\mathcal{D}_{2}\).
The resulting prediction interval for \(Y_{n+1}\) is then 
\begin{equation*}
  \left[\widehat{f}_{n_{1}}(X_{n+1})-{q}^{*}_{n_{2},\text{Pear}}\cdot\hat{\rho}_{n_1}(X_{n+1})\cdot[\widehat{f}_{n_{1}}(X_{n+1})]^{p/2},
  \widehat{f}_{n_{1}}(X_{n+1})+{q}^{*}_{n_{2},\text{Pear}}\cdot\hat{\rho}_{n_1}(X_{n+1})\cdot[\widehat{f}_{n_{1}}(X_{n+1})]^{p/2}\right].
\end{equation*}
It is important to note that \cite{lei2018distribution} highlighted that in linear regression, when computing residuals in $\mathcal{D}_{2}$, the prediction band width derived from $Y-X\hat{\betavec}$ exhibits minimal sensitivity to variations in X, provided the fitting procedure maintains reasonable stability. Nevertheless, this characteristic of approximately constant width across different covariate values becomes inadequate when the residual variance demonstrates substantial dependence on X, necessitating adaptive conformal intervals that can accommodate the underlying heteroscedastic nature of the errors.

In the case of Pearson residuals, however, the difference $Y-\widehat{f}_{n_{1}}(X)$ is already scaled by a function of $X$ through the variance function.  In this proposed local method, we allow for extra flexibility and local variability beyond the (possibly mis-specified) GLM mean-variance relationship.  
This approach essentially allows $\widehat{\phi}$, which was assumed constant previously for a true GLM and canceled out in the interval in Equation~\ref{eq:intp}, to vary with $X$.  

\begin{Remark}
While we may be fitting GLM-type models or using LightGBM with a GLM-inspired loss function (Tweedie), our goal is to construct prediction intervals without assuming that the GLM model specification is necessarily correct. The three residual types initially considered -- Pearson, deviance, and Anscombe -- are derived under the Tweedie model. However, if the model is severely misspecified, then these residuals, when used as nonconformity measures, offer no stronger theoretical justification than alternatives such as the locally weighted raw residuals proposed by \cite{lei2018distribution}, which target the raw residuals directly. The readers should note that conformal analysis works theoretically for any conformity scores that are exchangeable. Simulation results provided in Supplemental Materials Section 2 indicate that the locally weighted Pearson residuals and locally weighted residuals of
\cite{lei2018distribution} perform similarly, although the locally weighted residuals of \cite{lei2018distribution} tend to result in intervals with slightly higher than nominal coverage.  
\end{Remark}

\section{Illustration using Insurance Data}\label{simulation}

We consider the \textbf{AutoClaim} data from the \texttt{cplm} R package \citep{zhang2013cplm}. 
It contains 10296 records and 28 variables, described in Table~\ref{tab:autoclaim-data}. A more detailed description is provided in \cite{manna2024interval}. We take \texttt{CLM\_AMT5} as the dependent variable (\(Y\)), and all the other
variables except \texttt{POLICYNO}, \texttt{PLCYDATE}, \texttt{CLM\_FREQ5} and
\texttt{CLM\_AMT}, as the independent variables (\(X\)). In Figure 5 from \cite{manna2024interval}, there is a spike at zero in the distribution of claim amount. 
For a continuous variable with zero-inflation, the Tweedie regression model is one of the foremost models used in the insurance industry \citep[see][]{Halder_2023}.

\subsection{Simulation Study}

To construct and assess different split conformal prediction intervals, we repeatedly randomly partition the \textbf{AutoClaim} data
into training, calibration, and validation sets. We reserve a third partition of the data -- the validation set -- to compute the empirical coverage probabilities, which quantify how often the true response falls within the prediction intervals on the validation set. In practice, a practitioner interested in constructing prediction intervals would typically split the data into two equal parts: a proper training set and a calibration set. The size of the validation set is not crucial; its primary purpose is to assess the performance of the conformal prediction intervals for this article. The sets are defined as the proper training set \(\mathcal{D}_{1}\) (of size
\(n_{1}=4000\)), the calibration set \(\mathcal{D}_{2}\) (of size
\(n_{2}=4000\)), and the validation set \(\mathcal{D}_{3}\) (of size
\(n_{3}=2296\)). 

We considered regularized Tweedie GLM with elastic net penalty and LigthGBM with Tweedie loss to obtain $\hat{f}_{n_1}$, the estimated conditional mean function based on the data in $\mathcal{D}_1$.
Note that a regularized GLM minimizes the below objective:

\begin{equation*}
\min_{\beta_0, \beta} \frac{1}{n_1} \sum_{i=1}^{n_1}  l(y_i, \beta_0 + \beta^T x_i) + \lambda \left[ (1 - \gamma) \|\beta\|_2^2/2 + \gamma \|\beta\|_1 \right]
\end{equation*}
Here, $l(\cdot,\cdot)$ denotes a Tweedie negative log-likelihood. For the regularized GLM (referred to as GLMNET henceforth), we performed a grid search over elastic net mixing weights $\gamma \in  \{0.0, 0.1, 0.2, \ldots, 1.0\}$, where $0$ corresponds to ridge regression and $1$ corresponds to lasso. In addition, we considered Tweedie distributions with power parameters $p \in \{1.1, 1.2, \ldots, 1.9\}$. The optimal combination of $\gamma$, $p$, over the grid as mentioned above and the strength of elastic net penalty, $\lambda$, was selected by minimizing the cross-validation error using five-fold cross-validation. We used \texttt{glmnet} \citep{hastie2014glmnet} and \texttt{tweedie} \citep{dunn2017package} \texttt{R} packages for GLMNET.

LightGBM stands out among different tree-based methods due to its leaf-wise tree growth strategy, which differs from the level-wise growth used by XGBoost and the random subset approach of Random Forest. This leaf-wise approach enables LightGBM to focus on the most promising splits, offering the fastest training speed among the three (\cite{choudhury2024searches}), although it comes with a higher risk of overfitting. The leafwise growth helps to deal with the problem typical in imbalanced data (\cite{zhang2020comparison}). LightGBM naively supports categorical features, eliminating the need for pre-processing or label encoding required by XGBoost and Random Forest. It is also highly memory-efficient, making it suitable for large datasets where Random Forest might struggle with high memory usage. Furthermore, LightGBM handles missing values automatically, ensuring seamless data processing, and excels in distributed training scenarios due to its highly efficient parallelization. These features position LightGBM as a powerful and scalable option, especially for large-scale machine-learning tasks \citep[see, e.g., ][]{clemente2023modelling, wilson2024comparison}. Although both Gradient Boosting Models and GLMs can capture nonlinear interactions between predictors, GLMs achieve this through interaction terms and polynomial features, which significantly increases model complexity and computational burden. A detailed summary for LightGBM is obtained in \cite{ke2017lightgbm} or \url{https://lightgbm.readthedocs.io/en/latest/R/index.html}.

For the LightGBM model, we used the Tweedie objective with a power parameter ranging from 1.1 to 1.9. The number of leaves was fixed at 10, the learning rate was set to 0.005, and the maximum number of boosting iterations was 2000. The remaining parameters were used as the default parameters in the LightGBM package (see \cite{ke2017lightgbm}). The loss function was chosen as Tweedie loss in our experiment. We train the Tweedie model on the proper training set
\(\mathcal{D}_{1}\) using profile likelihood, in which given each \(p\), we
train  using gradient boosting \citep[LightGBM]{LightGBM_r_pkg} with cross-validation on tuning to obtain \(\widehat{f}_{n_1}(x;p)\). With
\(p\) and \(\widehat{f}(x;p)\), we obtain the dispersion estimate
\(\widehat{\phi}(p)\) using maximum likelihood estimation. We then compare the maximized likelihood for each triplet
(\(p,\widehat{f}(x;p),\widehat{\phi}(p)\)) on a sequence of choices of
\(p\in \{1.1, 1.2, \ldots, 1.9\}\)  to determine the optimal power parameter \(\widehat{p}\). Note that this training process is solely based on \(\mathcal{D}_{1}\). Five-fold cross-validation was used to select the optimal number of iterations and Tweedie power parameter that minimized the validation loss.

\begin{table}[!htbp]
\centering
\caption{Description of variables in the \textbf{AutoClaim} data}
\label{tab:autoclaim-data}
\resizebox{\textwidth}{!}{%
\begin{tabular}{||l|l|l||}
\hline
\textbf{Variable}   & \textbf{Type} & \textbf{Description}                                                                                                           \\ \hline\hline
\texttt{POLICYNO}   & character     & the policy number                                                                                                              \\
\texttt{PLCYDATE}   & date          & policy effective date                                                                                                          \\
\texttt{CLM\_FREQ5} & integer       & the number of claims in the past 5 years                                                                                       \\
\texttt{CLM\_AMT5}  & integer       & the total claim amount in the past 5 years                                                                                     \\
\texttt{CLM\_AMT}   & integer       & the claim amount in the current insured period                                                                                 \\
\texttt{KIDSDRIV}   & integer       & the number of driving children                                                                                                 \\
\texttt{TRAVTIME}   & integer       & the distance to work                                                                                                           \\
\texttt{CAR\_USE}   & factor        & the primary use of the vehicle: ``Commercial'', ``Private''                                                                    \\
\texttt{BLUEBOOK}   & integer       & the value of the vehicle                                                                                                       \\
\texttt{RETAINED}   & integer       & the number of years as a customer                                                                                              \\
\texttt{NPOLICY}    & integer       & the number of policies                                                                                                         \\
\texttt{CAR\_TYPE}  & factor        & the type of the car: ``Panel Truck'', ``Pickup'', ``Sedan'', ``Sports Car'', ``SUV'', ``Van''                                  \\
\texttt{RED\_CAR}   & factor        & whether the color of the car is red: ``no'', ``yes''                                                                           \\
\texttt{REVOLKED}   & factor        & whether the driver's license was revoked in the past 7 years: ``No'', ``Yes''                                                  \\
\texttt{MVR\_PTS}   & integer       & MVR violation records                                                                                                          \\
\texttt{CLM\_FLAG}  & factor        & whether a claim is reported: ``No'', ``Yes''                                                                                   \\
\texttt{AGE}        & integer       & the age of the driver                                                                                                          \\
\texttt{HOMEKIDS}   & integer       & the number of children                                                                                                         \\
\texttt{YOJ}        & integer       & years at current job                                                                                                           \\
\texttt{INCOME}     & integer       & annual income                                                                                                                  \\
\texttt{GENDER}     & factor        & the gender of the driver: ``F'', ``M''                                                                                         \\
\texttt{MARRIED}    & factor        & married or not: ``No'', ``Yes''                                                                                                \\
\texttt{PARENT1}    & factor        & single parent: ``No'', ``Yes''                                                                                                 \\
\texttt{JOBCLASS}   & factor        & ``Unknown'', ``Blue Collar'', ``Clerical'', ``Doctor'', ``Home Maker'', ``Lawyer'', ``Manager'', ``Professional'', ``Student'' \\
\texttt{MAX\_EDUC}  & factor        & max education level: ``<High School'', ``Bachelors'', ``High School'', ``Masters'', ``PhD''                                    \\
\texttt{HOME\_VAL}  & integer       & the value of the insured's home                                                                                                \\
\texttt{SAMEHOME}   & integer       & years in the current address                                                                                                   \\
\texttt{AREA}       & factor        & home/work area: ``Highly Rural'', ``Highly Urban'', ``Rural'', ``Urban''                                                       \\ \hline
\end{tabular}%
}
\end{table}


For computing the non-conformity quantile (as defined in Step 4 of Algorithm \ref{alg:glm}), we calculate, using the $\hat{f}_{n_1}$ obtained from both GLMNET and LightGBM, the following absolute residuals on the calibration set \(\mathcal{D}_{2}\): Pearson (Equation~\eqref{eq:resp}), Deviance (Equation~\eqref{eq:resd}), Anscombe (Equation~\eqref{eq:resa}), Unstandardized ($|Y_i-\hat{f}_{n_1}(X_i)|$), and Locally Weighted Pearson 
(Equation~\eqref{eq:respw}). These residuals are then used to construct prediction intervals as described in Algorithm~\ref{alg:glm} and Section~\ref{localized_formulation} for the dependent variable on the validation set \(\mathcal{D}_{3}\). Note that the prediction intervals based on Locally Weighted Pearson residuals also require a second model to be fitted to the Pearson Residuals in \(\mathcal{D}_{1}\) to obtain an estimate of the error spread, $\widehat{\rho}_{n_1}$ (as outlined in Section \ref{localized_formulation}). Note also that since choosing a penalized GLM as the second model for the error spread does not seem sensible, we implemented the Locally Weighted Pearson procedure only using LightGBM both for estimating $\hat{f}_{n_1}$ (using a Tweedie loss function) and $\widehat{\rho}_{n_1}$ (using a squared error loss).

\begin{figure}[ht]
\centering

\begin{subfigure}[b]{0.45\textwidth}
    \centering
    \includegraphics[width=\textwidth]{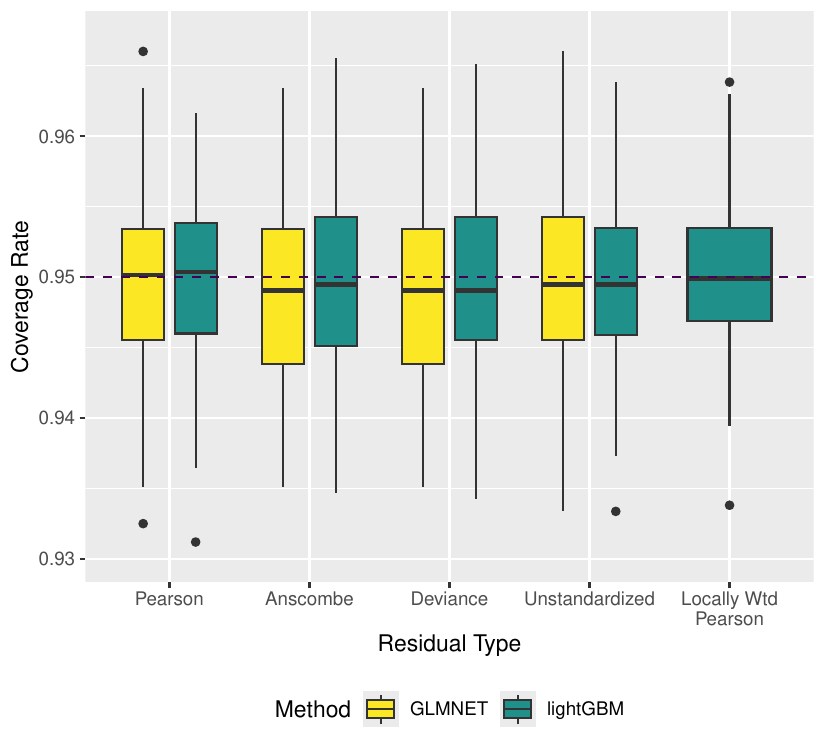}
    \caption{Coverage Rates}
    \label{fig:rates_symmetric_wo_lei}
\end{subfigure}
\hfill
\begin{subfigure}[b]{0.45\textwidth}
    \centering
    \includegraphics[width=\textwidth]{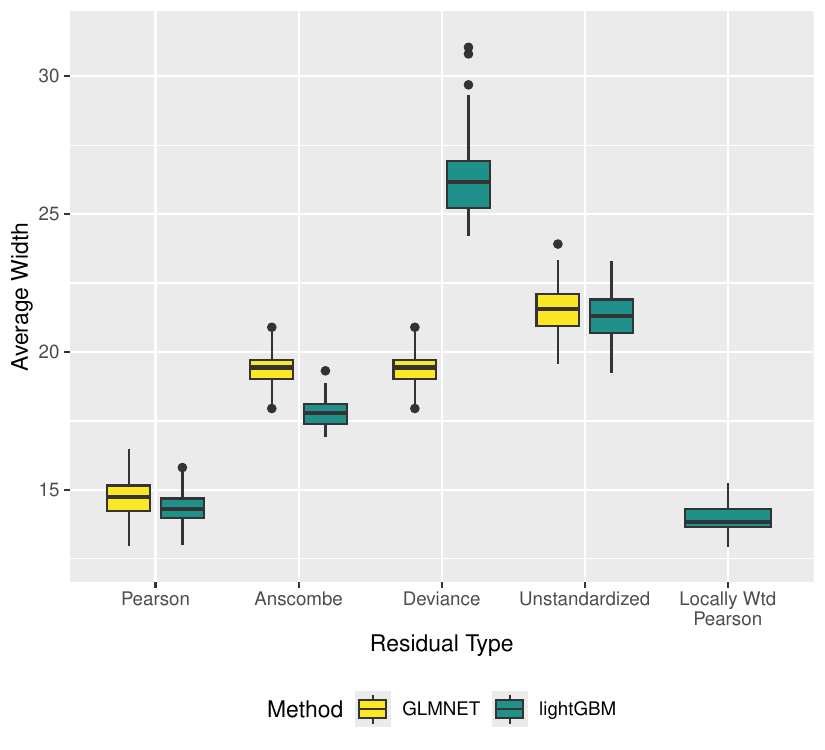}
    \caption{Average Widths}
    \label{fig:widths_symmetric_wo_lei}
\end{subfigure}

\caption{Results on conformal prediction intervals from 100 simulations using absolute residuals: (a) coverage rates and (b) average interval widths.}
\label{fig:combined_main_paper}
\end{figure}

We repeat the random partitioning of the \textbf{AutoClaim} data into \(\mathcal{D}_{1}\), \(\mathcal{D}_{2}\), and \(\mathcal{D}_{3}\) one hundred times and report the
coverage rates and average interval widths in \(\mathcal{D}_{3}\) across the 100 repetitions. 
The distribution of coverage rates and average widths of the prediction intervals over the 100 simulations are shown in Figures~\ref{fig:rates_symmetric_wo_lei} and~\ref{fig:widths_symmetric_wo_lei}. The mean average widths and coverage rates along with the standard deviations (SD) are reported in Table~\ref{tab:rates_widths_symmetric_wo_lei_tab}.
The summary statistics reported in Table~\ref{tab:rates_widths_symmetric_wo_lei_tab} are based on repeated evaluations over 100 random partitions. For each partition, prediction intervals are constructed for all 2,296 observations in the held-out test set $\mathcal{D}_3$, and the average width and empirical coverage rate are computed. The values reported in the table are the means and standard deviations (SD) of these 100 average widths and coverage rates, respectively. That is, each box plot in the figure summarizes the distribution of these 100 values across the different random splits. 
All residual types result in intervals achieving coverage close to the nominal level 0.95. Standard deviations are low (0.005–0.0066), indicating that interval performance across simulations is consistent. LightGBM generally produces narrower intervals than GLMNET, especially for Pearson and Anscombe residuals. The locally weighted Pearson residuals offer the narrowest intervals overall for LightGBM. Deviance residuals show a notable width inflation for LightGBM, suggesting potential instability, while unstandardized residuals yield the widest intervals.

For each of the 100 repetitions, we identified the best-performing GLMNET and LightGBM models, defined as those achieving the lowest cross-validated prediction error using hyperparameters selected
as previously described. In the case of GLMNET, the optimal value of elastic net mixing parameter was 1 (i.e., a lasso penalty) in 79 out of the 100 simulations. The best Tweedie power parameter, $p$, for GLMNET was consistently 1.5 across all 100 simulations. For LightGBM, the optimal $p$ was 1.3 in 87 of the 100 runs, and the average optimal number of boosting iterations was approximately 708.

\begin{table}[ht]
\centering
\caption{Mean ($\pm$ SD) Average Widths and Coverage Rates in the validation set ($\mathcal{D}_3$) of Conformal Prediction Intervals based on Absolute value of Residuals over 100 simulations}
\label{tab:rates_widths_symmetric_wo_lei_tab}
\resizebox{\textwidth}{!}{%
\begin{tabular}{@{}llcc@{}}
\toprule
\textbf{Residual Type}          & \textbf{Method} & \multicolumn{1}{l}{\textbf{Coverage (Mean ± SD)}} & \multicolumn{1}{l}{\textbf{Width (Mean ± SD)}} \\ \midrule
\multirow{2}{*}{Pearson}        & GLMNET          & 0.9495 ± 0.0061                                   & 14.76 ± 0.62                                   \\
                                & LightGBM        & 0.9496 ± 0.0057                                   & 14.32 ± 0.55                                   \\ \midrule
Locally Weighted Pearson        & LightGBM        & 0.9502 ± 0.0054                                   & 13.96 ± 0.51                                   \\ \midrule
\multirow{2}{*}{Anscombe}       & GLMNET          & 0.9487 ± 0.0062                                   & 19.35 ± 0.52                                   \\
                                & LightGBM        & 0.9495 ± 0.0059                                   & 17.79 ± 0.50                                   \\ \midrule
\multirow{2}{*}{Deviance}       & GLMNET          & 0.9487 ± 0.0062                                   & 19.35 ± 0.52                                   \\
                                & LightGBM        & 0.9496 ± 0.0060                                   & 26.32 ± 1.34                                   \\ \midrule
\multirow{2}{*}{Unstandardized} & GLMNET          & 0.9495 ± 0.0060                                   & 21.51 ± 0.89                                   \\
                                & LightGBM        & 0.9496 ± 0.0060                                   & 21.25 ± 0.92                                   \\ \bottomrule
\end{tabular}%
}
\end{table}

\subsection{Results from a Single Resampling}

We focus now on the analysis from a single random partion of the data into the three sets $\mathcal{D}_1$,  $\mathcal{D}_2$, and  $\mathcal{D}_3$, to discuss results and demonstrate the utility of resulting prediction intervals in the insurance industry.  
We observed that the nominal error rate is similar for both LightGBM and GLMNET; however, the prediction interval width, which measures the uncertainty, is lower for LightGBM. In the supplement (section 3), we also demonstrate that the predictive performance also shows that LightGBM performs better than GLMNET.
\begin{table}[ht]
\centering
\caption{LightGBM Feature Importance Scores by Gain, Cover, and Frequency}
\label{tab:feature-importance}
\begin{tabular}{@{}lrrr@{}}
\toprule
\textbf{Feature}   & \textbf{Gain} & \textbf{Cover} & \textbf{Frequency} \\ \midrule
\texttt{REVOLKED}  & 0.4492        & 0.2059         & 0.0816             \\
\texttt{MVR\_PTS}  & 0.2044        & 0.2396         & 0.1830             \\
\texttt{AREA}      & 0.1353        & 0.1928         & 0.1177             \\
\texttt{HOME\_VAL} & 0.0483        & 0.0693         & 0.1158             \\
\texttt{INCOME}    & 0.0246        & 0.0546         & 0.0790             \\
\texttt{TRAVTIME}  & 0.0212        & 0.0292         & 0.0656             \\
\texttt{BLUEBOOK}  & 0.0209        & 0.0282         & 0.0689             \\
\texttt{CAR\_TYPE} & 0.0171        & 0.0448         & 0.0478             \\
\texttt{JOBCLASS}  & 0.0168        & 0.0446         & 0.0486             \\
\texttt{YOJ}       & 0.0152        & 0.0178         & 0.0420             \\
\texttt{NPOLICY}   & 0.0135        & 0.0199         & 0.0399             \\
\texttt{AGE}       & 0.0112        & 0.0085         & 0.0371             \\
\texttt{SAMEHOME}  & 0.0091        & 0.0172         & 0.0338             \\
\texttt{KIDSDRIV}  & 0.0063        & 0.0121         & 0.0193             \\
\texttt{MAX\_EDUC} & 0.0034        & 0.0119         & 0.0088             \\
\texttt{HOMEKIDS}  & 0.0020        & 0.0006         & 0.0062             \\
\texttt{GENDER}    & 0.0007        & 0.0007         & 0.0020             \\
\texttt{CAR\_USE}  & 0.0006        & 0.0033         & 0.0027             \\
\texttt{MARRIED}   & 0.0000        & 0.0000         & 0.0002             \\ \bottomrule
\end{tabular}
\end{table}

Based on the best model selected in the LightGBM model in a single random resampling, we provide the matrix of feature importance in  Table~\ref{tab:feature-importance}. Gain represents the relative improvement in accuracy brought by a feature when it is used in decision trees; higher gain indicates important predictors. Cover measures the number of observations affected by the feature's splits, reflecting how broadly it applies across the data. Frequency refers to how often the feature is used across all trees in the model. In the context of the insurance industry, such insights help prioritize key risk factors -- for instance, the most influential variables in the model are \texttt{REVOLKED}, \texttt{MVR\_PTS}, and \texttt{AREA}, as they consistently exhibit the highest importance across all three metrics. In particular, \texttt{REVOLKED} stands out as the top predictor, indicating that whether a driver's license has been revoked plays a crucial role in determining the outcome. This is likely due to its direct relationship with driving risk or policy violations. Similarly, \texttt{MVR\_PTS} (motor vehicle record points) is a strong predictor, reflecting a driver’s history of infractions, which correlates with future risk. AREA also contributes significantly, suggesting regional patterns or socio-economic effects on behavior or claims.

Variables such as \texttt{HOME\_VAL}, \texttt{INCOME}, and \texttt{TRAVTIME} show moderate importance, pointing to the influence of economic factors and travel behavior. In contrast, demographic features like \texttt{GENDER}, \texttt{MARRIED}, and \texttt{HOMEKIDS} have minimal to no importance, implying limited predictive value. These results suggest that behavioral and geographic indicators are more predictive than static demographic attributes in this modeling context.

The feature importance analysis in Table~\ref{tab:feature-importance} provides valuable insights for insurance companies aiming to optimize underwriting, pricing, fraud detection, and customer targeting strategies. High-importance variables such as \texttt{REVOLKED} and \texttt{MVR\_PTS} (which capture license revocation status and motor vehicle record points) highlight the predictive value of driving history. Insurers can leverage these variables to more accurately assess driver risk and adjust premium pricing accordingly, thereby improving portfolio profitability.

Geographic and socioeconomic features such as \texttt{AREA}, \texttt{HOME\_VAL}, and \texttt{INCOME} also ranked highly, indicating that insurers can benefit from incorporating regional and demographic context into their risk models. For example, drivers in densely populated or high-traffic areas may present higher accident risk and could be priced differently or offered targeted safety incentives.

Additionally, features with low importance (e.g., \texttt{MARRIED}, \texttt{GENDER}, and \texttt{HOMEKIDS}) suggest opportunities for model simplification and bias mitigation. By de-emphasizing or excluding these variables, insurers may streamline models while aligning with evolving regulatory and ethical standards regarding fairness and transparency.

Hence, this information enables insurance companies to identify low-risk customers who may merit premium discounts, and high-risk individuals who require risk-adjusted pricing or more stringent underwriting. Stricter underwriting refers to the application of more rigorous evaluation criteria during risk assessment, helping insurers tailor coverage terms and manage potential liabilities more effectively.

\begin{figure}[htbp]
    \centering
    \includegraphics[width=0.9\textwidth]{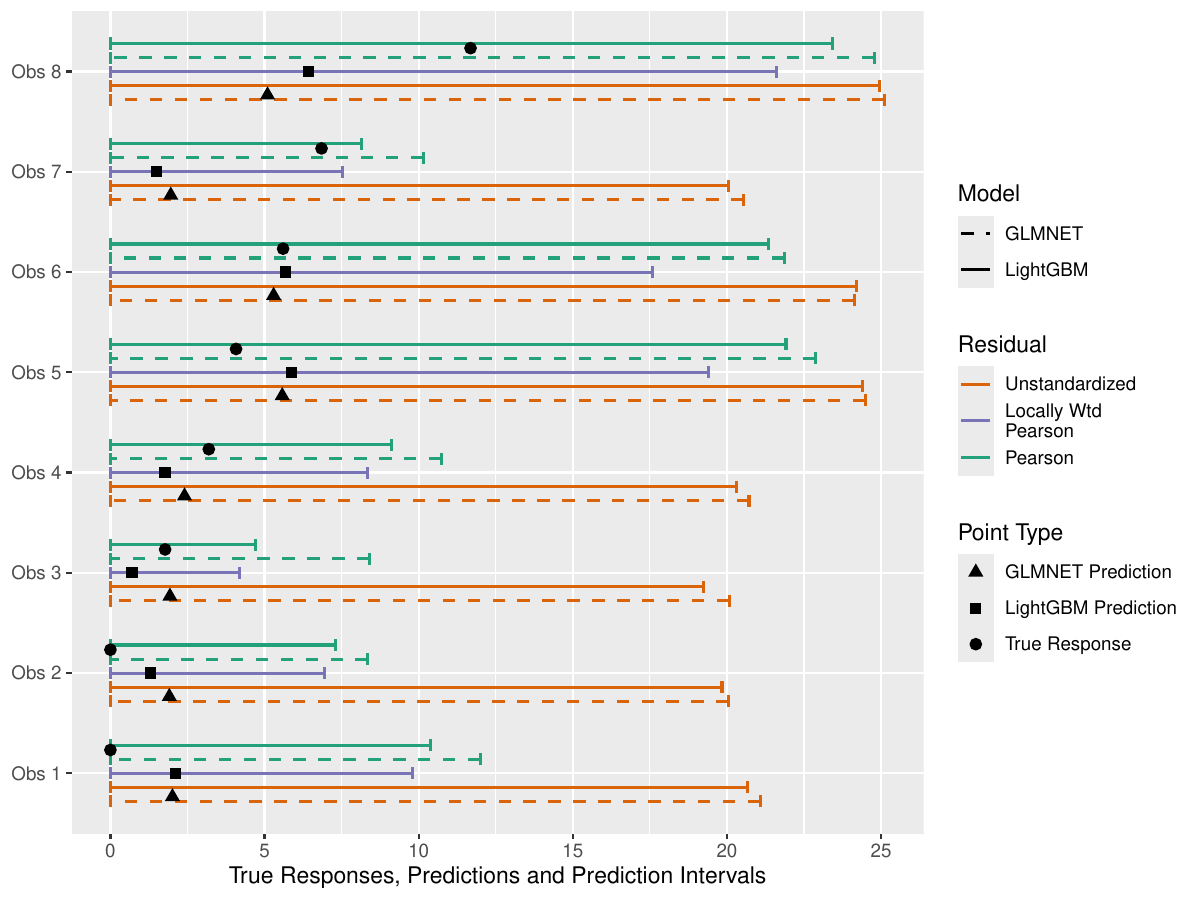}
    \caption{
       True Responses, Point Predictions and Prediction Intervals for Selected Observations}
    \label{fig:error_bars}
\end{figure}


The average width for LightGBM is smaller than other methods. We also investigated certain claims in $\mathcal{D_{3}}$. We selected eight claims, among which two claims are exactly zero, two are very extreme or large amounts of claims, and four are moderate level. 

Our results reveal that GLMNET consistently produces wider prediction intervals compared to LightGBM, indicating that LightGBM demonstrates superior uncertainty quantification capabilities. When examining Pearson residuals across different approaches, the localized weighted version yields the narrowest prediction intervals, suggesting enhanced precision in uncertainty estimation.

These prediction intervals serve a critical function in actuarial practice by quantifying the uncertainty inherent in individual forecasts. This uncertainty quantification enables actuaries and insurance professionals to make more informed decisions at the granular level of individual accounts and policies, moving beyond point estimates to embrace the full spectrum of potential outcomes.

Based on these findings, our proposed localized conformal prediction method using Pearson residuals with LightGBM emerges as an effective framework that simultaneously delivers accurate predictions and robust uncertainty quantification for insurance applications.


\section{Discussion}
\label{discussion}

In this work, we examined split conformal prediction inference for GLM-type models, introducing different types of known residuals such as Pearson, Deviance, and Anscombe residuals, as non-conformity measures. 
Focusing on auto claims and the Tweedie Compound Poisson-Gamma distribution for illustration, we show in empirical analysis that the locally weighted Pearson intervals with LightGBM provide the shortest intervals with the correct coverage.

We do not include Jackknife \citep{shao2012jackknife} 
in our simulations or data analysis, as its computational cost may be prohibitive for GLMs. \cite{dustin2024post} examines two bootstrap-based methods for constructing prediction intervals with post-selection inference, but concludes that the resulting intervals are too wide to be practically useful.  The key assumption for conformal prediction is the exchangeability of the (standardized) residuals, which underpins the validity of the predictive intervals. Therefore, the right-skewness of the Tweedie distribution should not, in itself, compromise performance - as long as the residuals are approximately exchangeable, we would expect symmetric intervals to perform well. We also explored constructing asymmetric intervals using raw residuals (rather than absolute residuals), but found no meaningful improvement over the symmetric intervals based on absolute residuals.



\section{Author Credit statement} 

In this paper, Alokesh Manna (\url{https://sites.google.com/view/alokesh-manna/home} and \url{https://statistics.uconn.edu/person/alokesh-manna/}) served as the lead author, contributing to the writing, conceptualization, methodology, validation, formal analysis, and investigation. Aditya Vikram Sett (\url{https://statistics.uconn.edu/person/aditya-vikram-sett/}) conducted simulation studies for comparing different methods. Dr. Yuwen Gu (\url{https://statistics.uconn.edu/yuwen-gu/}) helped conceptualize formally the residual for GLM and incorporate it into the LightGBM framework. Dr. Dipak Dey (\url{https://statistics.uconn.edu/person/dipak-dey/}) motivated us to compare the Bayesian methods and write the overall paper with his thoughtful suggestions. Professor \href{https://elizabeth-schifano.uconn.edu/}{Elizabeth Schifano (\url{https://elizabeth-schifano.uconn.edu/})} helped us with her comments on GLM and locally variant models, and also with the writing. Jichao He, Sr. Director \& Data Scientist at Travelers, (\url{https://www.linkedin.com/in/jichaohe/}) contributed to problem formulation across various real-world scenarios and practical applications of our approach.

\section{Acknowledgment}
\label{acknowledgment}
In this paper, we thank \href{https://sites.google.com/view/aritra-halder/home}{Aritra Halder (\url{https://sites.google.com/view/aritra-halder/home})} for his package \href{https://github.com/arh926/sptwdglm}{sptwdglm} to check with a Bayesian formulation. 
We thank the anonymous reviewers for their thoughtful comments and constructive suggestions, which have greatly helped improve the clarity and quality of this manuscript.

\section{Supporting Information}
We provide additional materials in the \ref{appendix}.

\clearpage 
\bibliographystyle{plainnat} 
\bibliography{lgb} 

\section{Appendix}
\label{appendix}

\section*{Raw residuals and Asymmetric Intervals}
\label{raw_residuals}

Although we focused on absolute residuals, one may consider the raw residuals to form asymmetric intervals. In that case, the definitions of the residuals are as follows:

\begin{align*}
    & R_{i,\text{Pear}} = \frac{Y_{i}-\hat{f}_{n_1}(X_{i})}
    {\widehat{\phi}^{1/2}[\hat{f}_{n_1}(X_{i})]^{p/2}}, \quad i \in \mathcal{D}_2, \\
    & R_{i,\text{Dev}} = \operatorname{sgn}(Y_{i}-\widehat{f}_{n_{1}}(\bX_{i}))
    \sqrt{2\left(\frac{Y_{i}[\widehat{f}_{n_{1}}(\bX_{i})]^{1-p}}{p-1}
      -\frac{Y_{i}^{2-p}}{(p-1)(2-p)}
      +\frac{[\widehat{f}_{n_{1}}(\bX_{i})]^{2-p}}{2-p}\right)}, \quad i \in \mathcal{D}_{2}, \\
   & R_{i,\text{Ansc}} = \frac{\frac{3}{3-p}(Y_{i}^{1-p/3}-[\widehat{f}_{n_{1}}(\bX_{i})]^{1-p/3})}
    {[\widehat{f}_{n_{1}}(\bX_{i})]^{p/6}}, \quad i \in \mathcal{D}_{2}.
\end{align*}
Note that in the definition of raw residuals, the residuals can be negative. For the convenience of our explanation, in general, we refer to the residuals as \(R_i\).
Let \( q_L \) and \( q_R \) denote the  
\(\lfloor \tfrac{\alpha}{2}(n_2 + 1) \rfloor\)th and  
\(\lceil (1 - \tfrac{\alpha}{2})(n_2 + 1) \rceil\)th  
smallest values of \( R_i \), for \( i \in \mathcal{D}_2 \), respectively. That is,
\[
q_L = R_{\left(\lfloor \tfrac{\alpha}{2}(n_2 + 1) \rfloor \right)}, \quad
q_R = R_{\left(\lceil (1 - \tfrac{\alpha}{2})(n_2 + 1) \rceil \right)}.
\]
By exchangeability, the conformal prediction interval satisfies
\begin{equation}\label{eq:pi-asym}
\Pr\left(q_L \leq R_{n+1} \leq q_R\right) \geq 1 - \alpha.
\end{equation}
From Equation~\eqref{eq:pi-asym}, one can follow the same procedure for absolute residuals to obtain the prediction interval for \(Y_{n+1}\) based on the Pearson and Anscombe residuals. We call the resulting interval the asymmetric prediction interval. 


\begin{table}[ht]
\centering
\caption{Average Coverage (± SD) for Asymmetric PIs with raw residuals over 100 Simulations}
\label{tab:coverage_table_asym}
\begin{tabular}{lll}
\toprule
\textbf{Residual Type} & \textbf{Method} & \textbf{Coverage (Mean ± SD)} \\
\midrule
\multirow{2}{*}{Pearson}        & GLMNET   & 0.9489 ± 0.0061 \\
                                & LightGBM & 0.9495 ± 0.0058 \\
\toprule                               
Locally Wtd Pearson (A variant) & LightGBM & 0.9499 ± 0.0066 \\
\toprule
\multirow{2}{*}{Anscombe}       & GLMNET   & 0.9492 ± 0.0062 \\
                                & LightGBM & 0.9497 ± 0.0058 \\
\toprule
\multirow{2}{*}{Deviance}       & GLMNET   & 0.9492 ± 0.0062 \\
                                & LightGBM & 0.9498 ± 0.0058 \\
\toprule
\multirow{2}{*}{Unstandardized} & GLMNET   & 0.9493 ± 0.0059 \\
                                & LightGBM & 0.9502 ± 0.0056 \\
\bottomrule
\end{tabular}
\end{table}

\begin{table}[ht]
\centering
\caption{Average Interval Width (± SD) for Asymmetric PIs with raw residuals over 100 Simulations}
\label{tab:width_table_asym}
\begin{tabular}{lll}
\toprule
\textbf{Residual Type} & \textbf{Method} & \textbf{Width (Mean ± SD)} \\
\midrule
\multirow{2}{*}{Pearson}        & GLMNET   & 22.19 ± 1.80 \\
                                & LightGBM & 22.18 ± 1.26 \\
 \toprule                               
Locally Wtd Pearson (A variant)  & LightGBM & 22.96 ± 1.86 \\
\toprule
\multirow{2}{*}{Anscombe}       & GLMNET   & 22.06 ± 1.72 \\
                                & LightGBM & 22.24 ± 1.22 \\
\toprule                                
\multirow{2}{*}{Deviance}       & GLMNET   & 22.06 ± 1.72 \\
                                & LightGBM & Computationally expensive \\
\toprule                                
\multirow{2}{*}{Unstandardized} & GLMNET   & 28.99 ± 1.14 \\
                                & LightGBM & 29.13 ± 0.92 \\
\bottomrule
\end{tabular}
\end{table}

In insurance analytics, obtaining valid prediction intervals is an 
essential component of understanding claim amounts. Although we are not focusing on the time series aspect of our model, the readers should note that there are several ongoing research to focus on the stochastic aspect in insurance modeling.  \cite{pires2022forecasting} 
described a time series model for forecasting home insurance amounts in different years using the SARIMA model and provided bootstrap and normality-based confidence 
intervals of the estimated value. Stochastic reserve models and uncertainty 
quantification are famous in insurance data claims (see \cite{wuthrich2008stochastic}). 
\cite{england2002stochastic} explored different stochastic reserve models, and 
Bornhuetter-Ferguson techniques in a Bayesian framework, and described different 
variabilities in claim reserve estimates. \cite{mohamed2024size} defined a new 
size-of-loss distribution for the negatively skewed insurance claims data and 
risk analysis related to the number of claims. \cite{omari2018modeling} discussed 
different statistical distributions of claim amounts and model selection criteria 
in auto insurance claims.





\begin{figure}[ht]
 \centering
 \begin{subfigure}[t]{\textwidth}
   \centering
   \includegraphics[width=\textwidth,height=7.5cm]{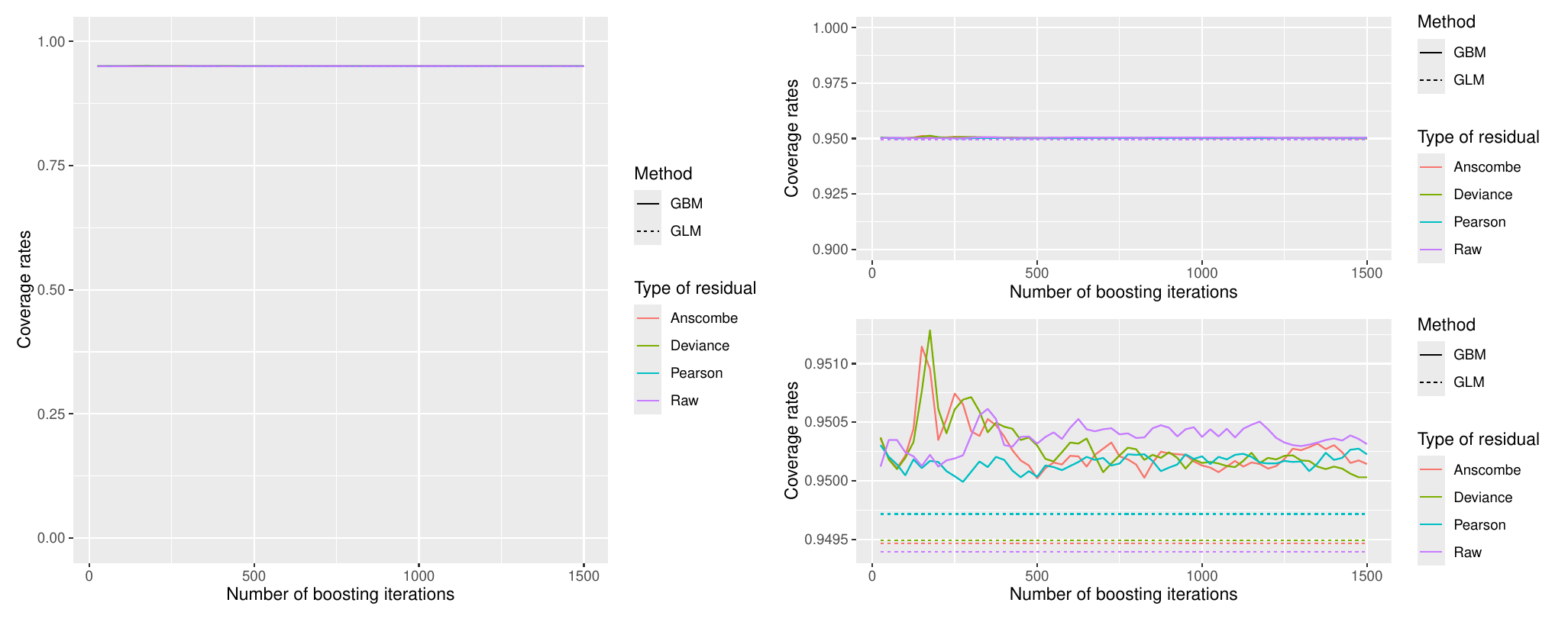}
   \caption{Average coverage rates (symmetric)}
   \label{fig:PI-coverage-sym}
 \end{subfigure}
 \hfill
 \begin{subfigure}[t]{\textwidth}
   \centering
   \includegraphics[width=\textwidth,height=7.5cm]{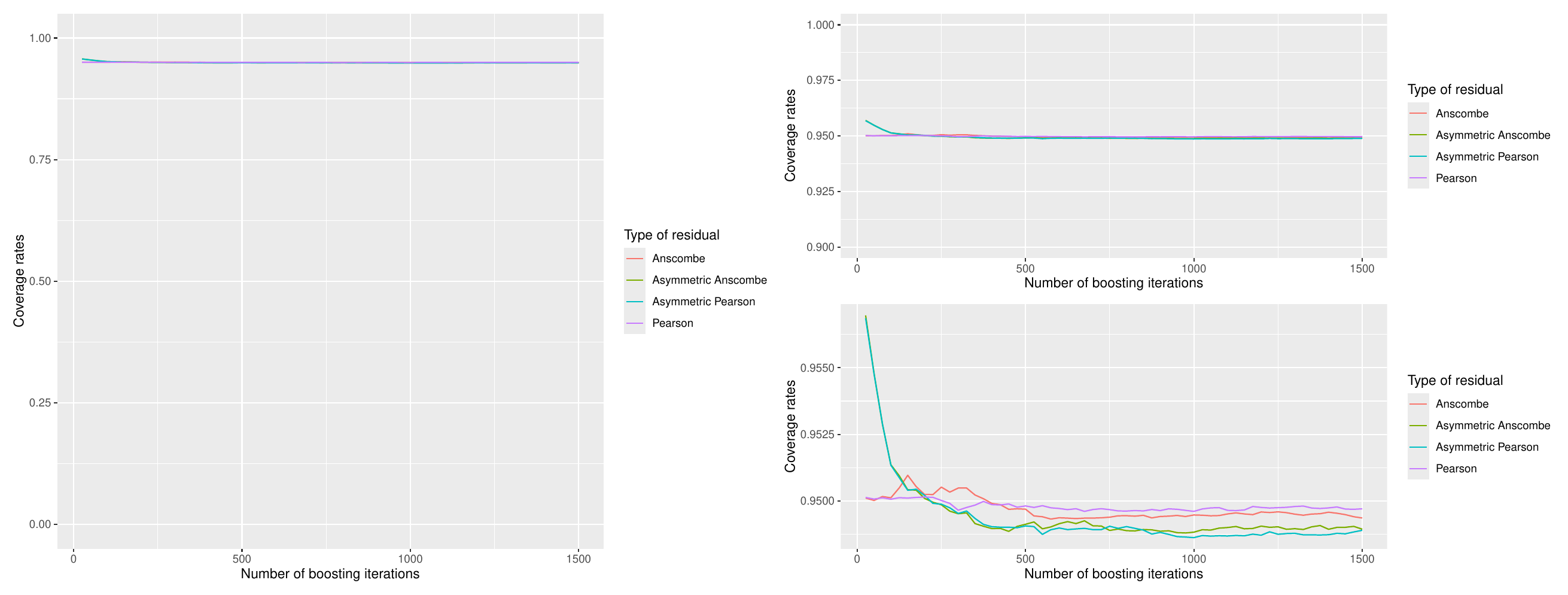}
   \caption{Average coverage rates (asymmetric)}
   \label{fig:PI-coverage-asym}
 \end{subfigure}
 \caption{Comparison of average coverage rates between symmetric and asymmetric cases.}
 \label{fig:PI-coverage-comparison}
\end{figure}

The average coverage rates across different boosting iterations are described in Figure~\ref{fig:PI-coverage-comparison}. The three panels in this figure show the importance of closer looking around the .95 boundary. The average length of the prediction interval is described in Figure~\ref{fig:PI-length-combined}. 

LightGBM has built-in support for handling categorical variables efficiently, one of its key strengths. For the traditional linear model, if the categorical variable has many unique categories (high cardinality), one-hot encoding can significantly increase the number of features, leading to computational inefficiency and overfitting risks. LightGBM offers a powerful, efficient way to incorporate these features without extensive pre-processing. The superior performance with respect to the regular GLM relies on the fact of better fitting capturing complex, nonlinear relationships in the data automatically through decision trees, making it better suited for datasets where the true relationships are not purely linear.

\section*{Additional Simulation Results}

The same procedure for resampling the auto claims data discussed in Section 3 of the main text was used to examine the performance of both asymmetric intervals, as described in the previous section, and the locally weighted approach of \cite{lei2018distribution}.

\begin{figure}[ht]
\centering
\begin{subfigure}[t]{0.45\textwidth}
    \centering
    \includegraphics[width=\textwidth]{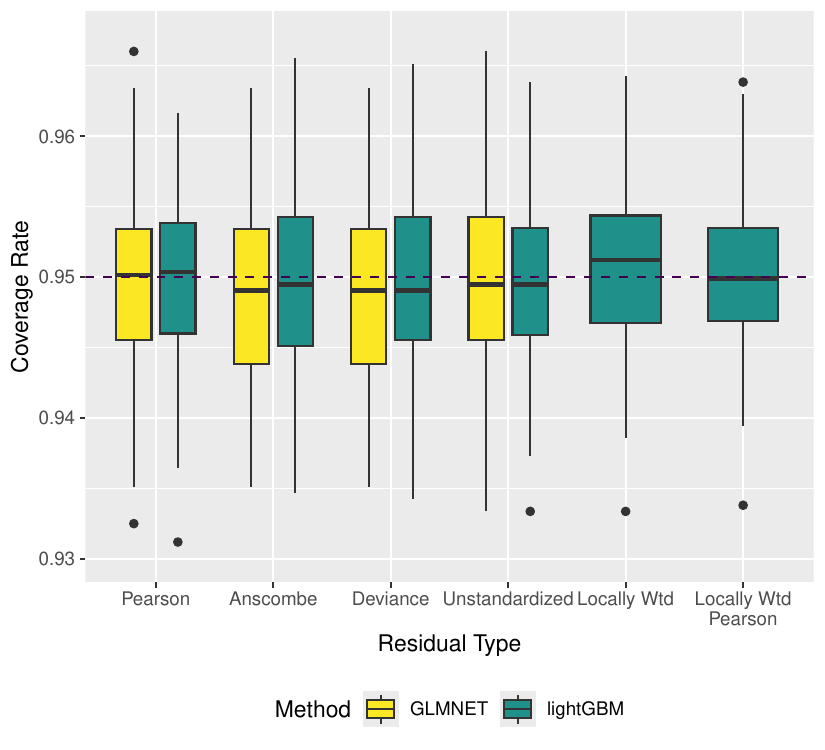}
    \caption{Coverage Rates of Symmetric Conformal PIs with Absolute Residuals}
    \label{fig:sub1_rates_sym_abs_resid}
\end{subfigure}
\hfill
\begin{subfigure}[t]{0.45\textwidth}
    \centering
    \includegraphics[width=\textwidth]{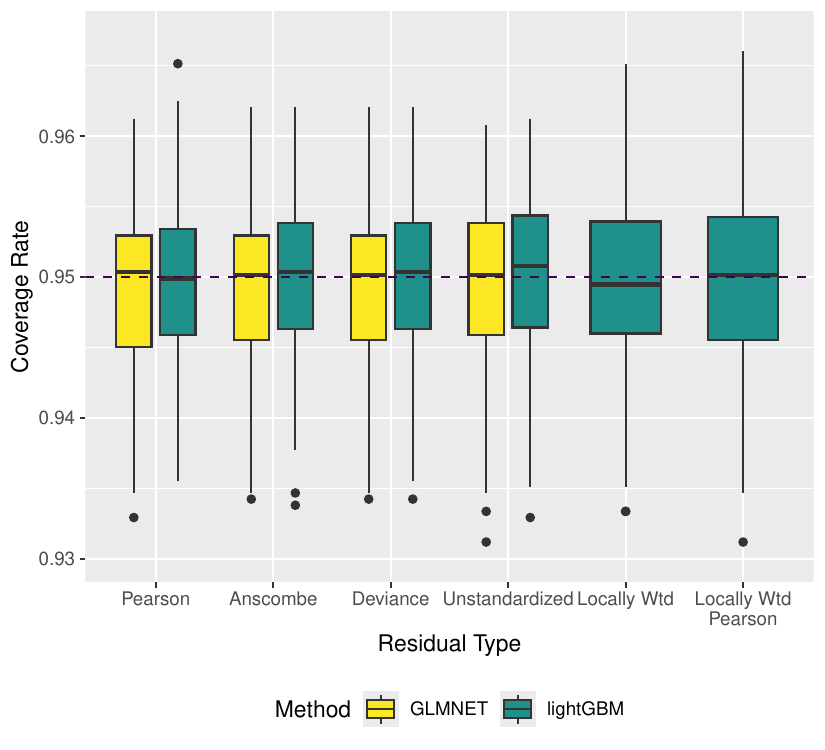}
    \caption{Coverage Rates of Asymmetric Conformal PIs with Raw Residuals}
    \label{fig:sub2_rates_asym_raw_resid}
\end{subfigure}

\vspace{1em}

\begin{subfigure}[t]{0.45\textwidth}
    \centering
    \includegraphics[width=\textwidth]{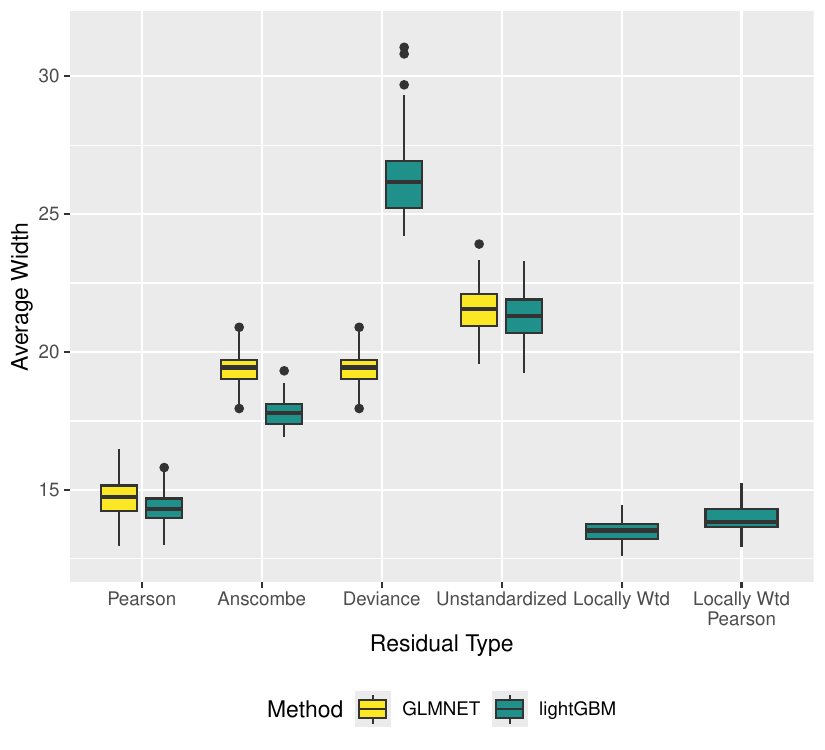}
    \caption{Average widths of Symmetric Conformal PIs with Absolute Residuals}
    \label{fig:sub3_widths_sym_abs_resid}
\end{subfigure}
\hfill
\begin{subfigure}[t]{0.45\textwidth}
    \centering
    \includegraphics[width=\textwidth]{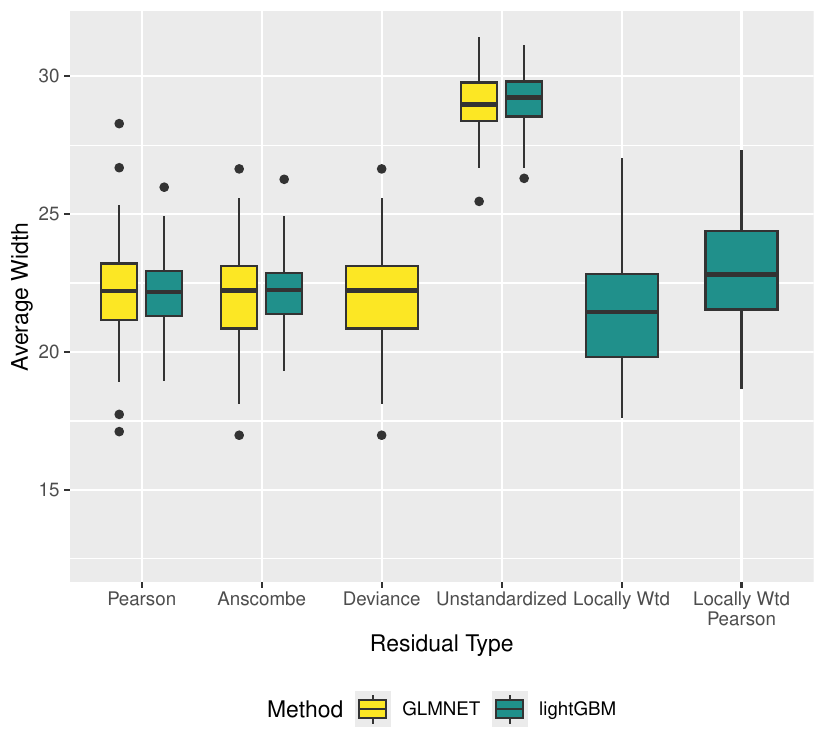}
    \caption{Average widths of Asymmetric Conformal PIs with Raw Residuals}
    \label{fig:sub4_widths_sym_raw_resid}
\end{subfigure}

\caption{Coverage Rates and Average Widths of Conformal PIs across 100 simulations}
\label{fig:combined_supplement}
\end{figure}

We can see in Figure~\ref{fig:combined_supplement} that while the asymmetric intervals have approximately correct coverage, the interval widths are much larger for the asymmetric intervals than the corresponding symmetric intervals.  Note for the Deviance residuals, the root finding procedure is computationally intensive and not very helpful for the practitioners, so we did not report the average interval width for the asymmetric deviance residuals. We additionally see the locally weighted Pearson residuals and locally weighted residuals of
\cite{lei2018distribution} perform similarly, although the locally weighted residuals of \cite{lei2018distribution} tend to result in symmetric intervals with slightly higher than nominal coverage.   

\section*{LightGBM and GLMNET comparison}

Based on our computation, we obtained the best parameter from $\mathcal{D}_1$ and used that for obtaining the performance comparison over the set $\mathcal{D}_3$. We see that in terms of prediction LightGBM performs better than GLMNET.

\begin{table}[ht]
\centering
\caption{Prediction Metrics in Validation Set $\mathcal{D}_3$, for one round of simulation}
\label{tab:metrics}
\begin{tabular}{@{}lrrrr@{}}
\toprule
Model    & RMSE  & MAE   & $R^2$ \\ \midrule
LightGBM & 7.399 & 4.005 & 0.283  \\
GLMNET   & 7.483 & 4.165 & 0.267  \\ \bottomrule
\end{tabular}
\end{table}



\end{document}